\documentclass[conference]{IEEEtran}

\usepackage{amsmath,amsfonts}
\usepackage{algorithm}
\usepackage{array}
\usepackage[caption=false,font=normalsize,labelfont=sf,textfont=sf]{subfig}
\usepackage{textcomp}
\usepackage{stfloats}
\usepackage{url}
\usepackage{verbatim}
\usepackage{graphicx}
\usepackage{cite}
\usepackage{tikz}
\usetikzlibrary{matrix,decorations.pathreplacing}
\IEEEoverridecommandlockouts
\IEEEpubid{\makebox[\columnwidth]{978-1-5386-5541-2/18/\$31.00~\copyright2024 IEEE \hfill}
\hspace{\columnsep}\makebox[\columnwidth]{ }}

\usepackage{times} 
\usepackage{float}
\usepackage{caption}
\usepackage{tabularx,ragged2e,booktabs}
\usepackage{url}
\usepackage{todonotes}
\usepackage{algpseudocode}
\usepackage{graphicx}

\usepackage[font=footnotesize]{caption}

\title{\LARGE \bf
Visuo-Tactile based Predictive Cross Modal Perception \\
for Object Exploration in Robotics
}
\author{Anirvan Dutta, Etienne Burdet and Mohsen Kaboli
\thanks{A Dutta and M Kaboli are with the BMW Group, RoboTac Lab, Munich, Germany. 
e-mail: name.surname@bmwgroup.com}%
\thanks{A Dutta and E Burdet are with Imperial College of Science, Technology and Medicine, London, UK. M Kaboli is with Eindhoven University of Technology, Netherlands.}
\thanks{Funded in part by the EU H2020 INTUITIVE under Grant ID 861166 and in part by EU Horizon PHASTRAC under Grant ID 101092096.}
}

\begin{document}
\bstctlcite{IEEEexample:BSTcontrol}
\maketitle
\IEEEpubidadjcol

\begin{abstract}
Autonomously exploring the unknown physical properties of novel objects such as stiffness, mass, center of mass, friction coefficient, and shape is crucial for autonomous robotic systems operating continuously in unstructured environments. We introduce a novel visuo-tactile based predictive cross-modal perception framework where initial visual observations (shape) aid in obtaining an initial prior over the object properties (mass). The initial prior improves the efficiency of the object property estimation, which is autonomously inferred via interactive non-prehensile pushing and using a dual filtering approach. The inferred properties are then used to enhance the predictive capability of the cross-modal function efficiently by using a human-inspired `surprise' formulation. We evaluated our proposed framework in the real-robotic scenario, demonstrating superior performance.  

\end{abstract}

\section{Introduction}
\label{sec:introduction}
\begin{figure}[t!]
    \centering
    \includegraphics[width=0.9\columnwidth]{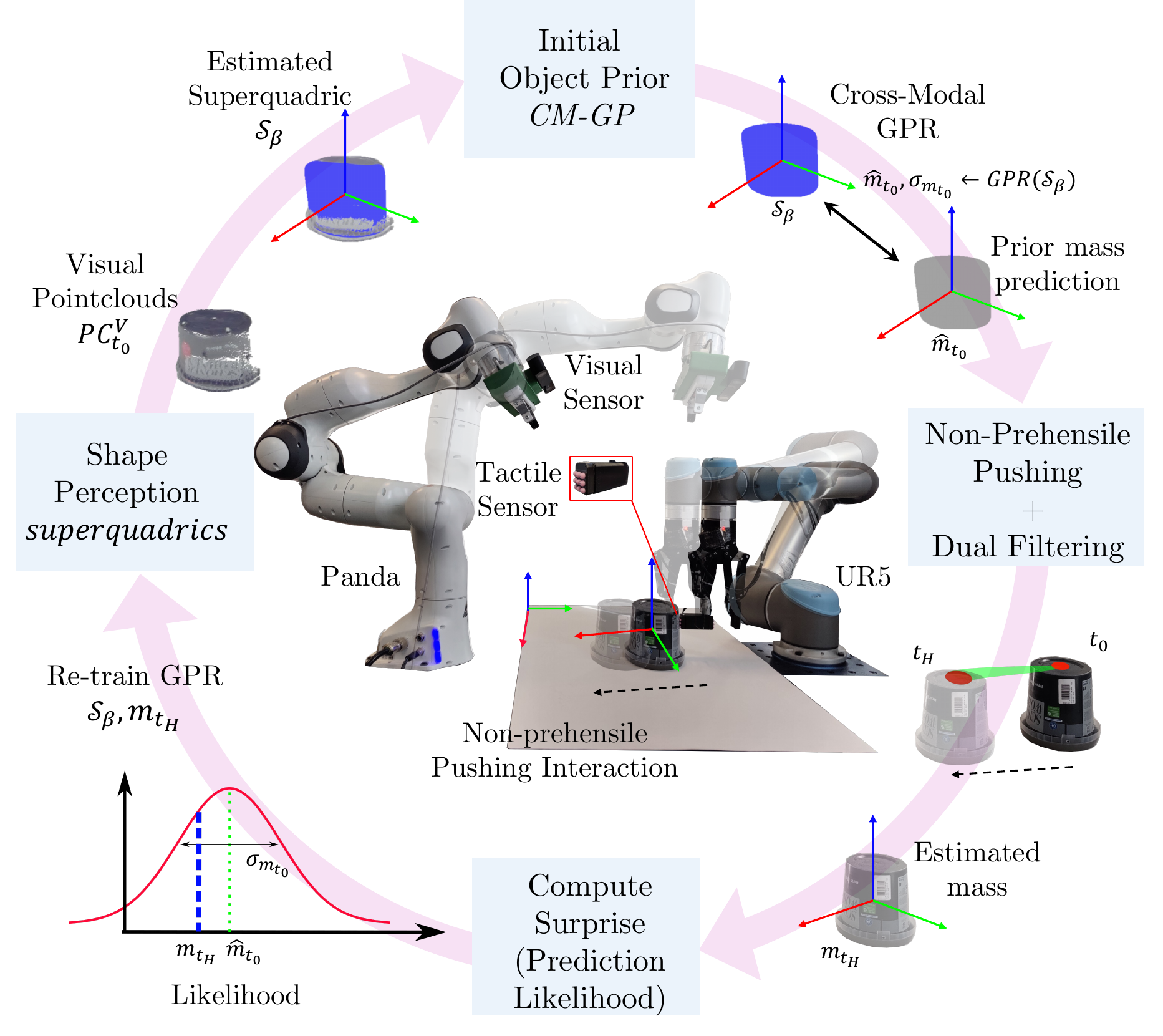}
    \caption{Problem setup for visuo-tactile based predictive cross-modal perception for object exploration}
    \label{fig:problem_setup}
\end{figure}

In order to handle novel objects, humans frequently rely on exploratory and interactive actions, such as holding, grasping, or pushing, to infer object properties such as shape, inertial, surface texture, and viscoelasticity \cite{etienne_3, kaboli2017tactile, kaboli2015humanoids, kaboli2019auro, kaboli_texture_tro}, which is formalized as  \textit{interactive visuo-tactile perception} \cite{bohg_17_interactive, bajcsy_active, seminara_active, kaboli_review_tro, murali2022active}. This study aims to enhance the inference of object properties during the visuo-tactile based robotic interaction by incorporating a predictive cross-modal prior over the object properties (see Fig.\ref{fig:problem_setup}).

We employ non-prehensile pushing to interact with objects and estimate/infer its properties \cite{mason_pushing}.  Unlike grasping and lifting \cite{kaboli2015hand, kaboli2016tactile}, pushing offers versatility, can be applied to objects of different sizes, and is also easier to perform. Furthermore, it does not require considerations like precise object geometry or grasp stability. Moreover, since the object is not tightly held by the robotic end effector during pushing, it allows a wider range of movements, which is helpful for property estimation \cite{mason1999progress}.

Nevertheless, using non-prehensile pushing to infer object parameters is a challenging task as the interaction dynamics between the object and the robotic system are sophisticated to model, due to uncertainty in the contacts, surface irregularities, etc.  In our previous work \cite{dutta2023push}, we proposed a novel dual differentiable filtering approach to consistently infer the physical properties of objects via non-prehensile pushing using a combination of visual and tactile sensing. However, in \cite{dutta2023push} we did not take advantage of any prior information about the object from the initial visual glance (cross-modal) to improve the inference process, which is addressed here.

Visuo-tactile-based cross-modal perception and learning is a challenging problem in robotics as there is a) a weak correlation between tactile and vision sensing \cite{murali2022deep}, b) the properties of physical objects are not salient under static or quasi-static interactions, and often each parameter is revealed only under specific interactions \cite{dense_phys}, and c) improving the cross-modal prediction requires a closed-loop interaction process. These challenges have limited the scope of previous research work. Previous works such as \cite{falco2019transfer, murali2022deep, parsons2022visuo} addressed cross-modal object recognition, mainly on static objects, thus limiting to shape recognition from tactile touches using prior visual knowledge. Other works \cite{yuan2017connecting, liu2018surface, zheng2019cross, lee2019touching} focused on the cross-modal synthesis of visual and tactile observations for textural properties. These works relied on training on a large dataset of isolated visual and tactile data, and the focus was on domain adaptation from visual to tactile or vice versa. However, the cross-modal synthesis of visual or tactile observation is a more challenging problem due to the intricate nature of tactile observations \cite{murali2023touch}, which are action-conditioned and therefore lack generalizability. In this regard, the cross-modal transfer between the properties (such as shape to mass, colour to texture, etc.) of an object represents a robust approach \cite{tatiya2023transferring}. 

The authors of \cite{martin2017cross} presented a cross-modal interactive perception approach for articulated objects. However, the approach was limited to estimating kinematic properties such as the pose of objects from proprioception in the case of visual occlusion. More recently, the authors in \cite{zheng2020lifelong, xiong2022deeply} proposed a lifelong visuo-tactile cross-modal learning framework for robotics. In this work, the authors proposed a common encoding space of visual and tactile information and used correlation to transfer information from one modality to another. The authors highlighted the issues faced for lifelong learning of the cross-modal function, such as catastrophic forgetting and overconfident prior estimates in the case of novel scenarios. Furthermore, the framework was limited to isolated visual and tactile interactions, and both modalities were not explicitly coupled in a closed loop. 

To address the above-mentioned problems, we take inspiration from the cross-modal perceptual processes observed in humans \cite{cm_nature}, in which prior visual knowledge about an object is seamlessly integrated before engaging in interactive perception. For example, based on visual cues, such as the size of the object, humans derive an initial estimate of its mass that regulates subsequent interactions. Moreover, the discrepancies between the initial prior and the actual physical properties estimated after interaction are akin to the `surprise' for humans, which also contributes to refining the prior prediction over time. This behavior has recently been formulated under the theory of predictive processing in computational neuroscience \cite{fep} and has served as the basis of our proposed framework. 

We propose a novel predictive cross-modal exploration framework in which initial visual observations (shape) aid in obtaining an initial prior over the object properties (mass) via a novel cross-modal function. Thereby the properties of the object are autonomously inferred or estimated via interactive non-prehensile pushing and using a dual filtering approach. The inferred properties are then used to enhance the predictive capability of the cross-modal function efficiently by using a human-inspired `surprise' formulation. Our framework allows the robotic system to interact and acquire novel information in an unsupervised fashion, facilitating lifelong cross-modal perception and learning. Our proposed framework is illustrated in Fig.\ref{fig:framework}.  

\subsection*{Contributions}
The contributions of the work are summarized in three folds.

\begin{enumerate}
    \item We propose a novel visuo-tactile cross-modal function (\textit{CM-GP}) based on Gaussian Process Regression to obtain a prior over the key property of the object of mass from a low-dimensional shape representation of an object along with associated uncertainty.
    \item We improve the efficiency of our dual filtering (DF) approach using the prior distribution over the mass provided by \textit{CM-GP}.
    \item We propose a novel human-inspired `surprise' formulation for efficient and long-term training of the Gaussian process function \textit{CM-GP} by the consistent estimate of object mass from interactive non-prehensile pushing and using our novel dual filtering approach.
\end{enumerate}
We performed extensive real-robot experiments that validated the efficacy of the proposed method compared to an approach without using cross-modal functionality. 
\section{Proposed Method}
\begin{figure*}[t!]
    \centering
    \includegraphics[width = 0.9\textwidth]{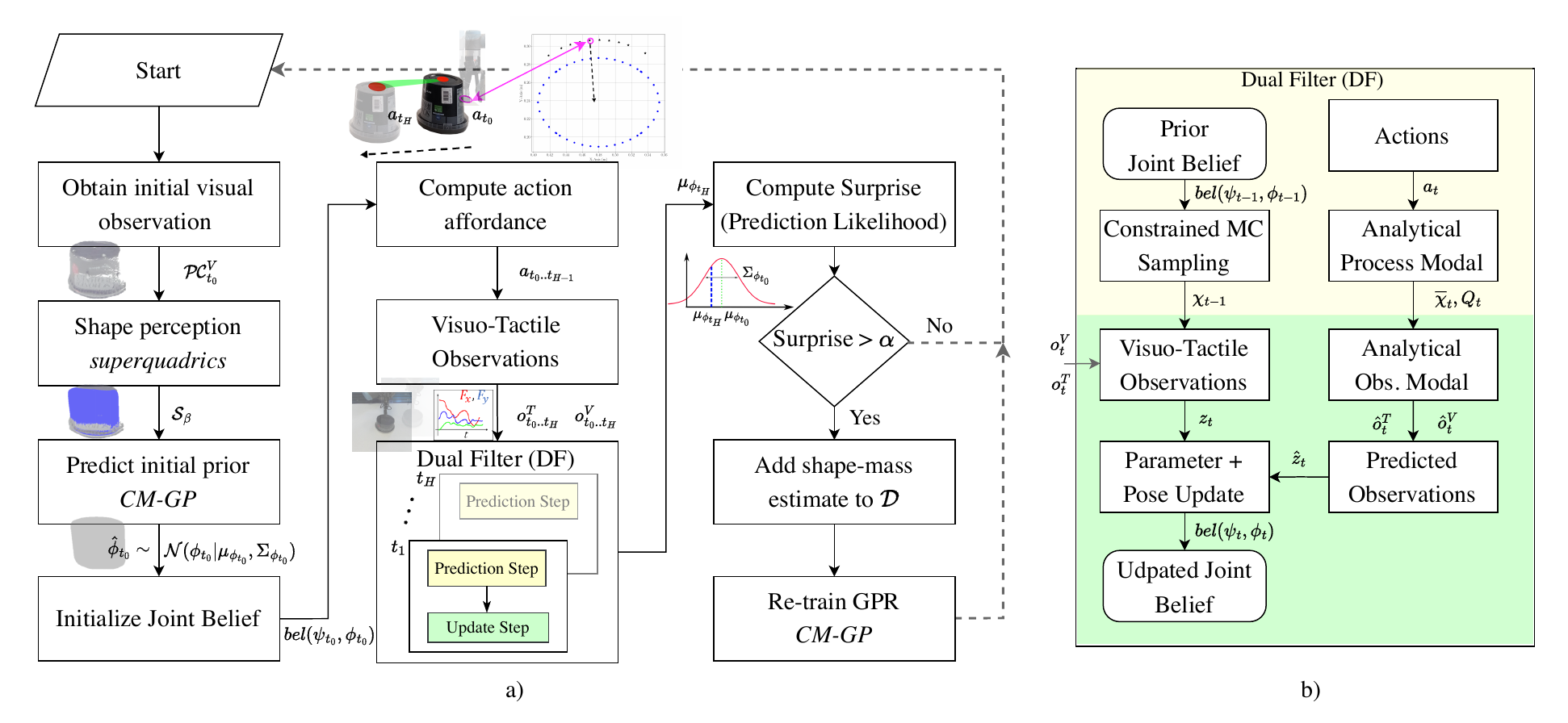}
    \caption{Our proposed framework (a) for visuo-tactile based predictive cross-modal perception of object properties. The detailed block of the dual filtering approach in (b).}
    \label{fig:framework}
\end{figure*}

\subsection{Problem Definition}
\label{sec:prob}
In this work, we consider the problem of estimating the state $s$  of an unknown rigid object from vision $o^V$ and tactile observation $o^T$ using non-prehensile pushing actions $a$. At any given time $t \in t_0, .. t_H$, the state $s_t = \{ \psi_t, \phi \}$ comprises of time-varying factors: \textbf{pose}, $\psi_t=\{x_t, y_t, \theta_t\}$ as well as object physical \textbf{parameter} of mass $\phi= \{m_t\}$. The observation $o^V_t$ consists of RGB-D images of the pushing area and tactile observation $o^T_t$ comprising of \textit{2D contact forces}. The pushing action $a_t$ is parameterized by \textit{contact point $(cp)$}, \textit{push direction} $(pd)$ and \textit{velocity} $(u)$ of the push. $cp$ consists of the 2D world coordinate of the contact point, $pd$, the z-axis rotational angle of the robotic system aligned along a pushing direction \& $u$ is the magnitude of the push velocity.

We perform quasi-static pushing \cite{mason_pushing} to infer the object state $\psi, \phi$, which is not directly observable through vision or tactile sensing. We leverage our novel dual filtering approach for the inference problem. Furthermore, we improve our inference by obtaining an initial prior distribution over the parameter of the object $\hat{\phi}_{t_0} \sim \mathcal{N}(.| \mu_\phi, \sigma_\phi)$ from vision using a cross-modal estimator \textit{CM-GP} which is approximated using the Gaussian process regression model (GPR) \cite{seeger2004gaussian}. The \textit{CM-GP} takes an initial low-dimensional shape estimate of the object using superquadrics $\mathcal{S}_\beta$ as input. The \textit{CM-GP} model is trained iteratively and uses only sparse pairwise datapoints of estimated shape and mass values $[X=\mathcal{S}_\beta, y=\phi_{t_H}] \in \mathcal{D}$. The dataset $\mathcal{D}$ is incrementally populated in an unsupervised manner using a novel `surprise' formulation making it viable for long-term use. Fig.\ref{fig:framework} presents an overview of our proposed framework, the details of which are presented in the following sections. 

\subsection{Superquadrics Shape Perception}
\label{sec:shape}
As a first step in exploring the state of an object, the shape of the object $\mathcal{S}$ is estimated using superquadrics, which are a family of geometric primitives. Superquadrics offer a rich shape vocabulary, such as cuboids, cylinders, ellipsoids, octahedra, and their intermediates, encoded by only five parameters. A superquadric centred in the origin with a frame aligned with the global x, y, z co-ordinate follows the following implicit function \cite{barr_sq}:
\begin{align}
    F(x, y, z) = {\left(\left(\frac{x}{a_x}\right)^{\frac{2}{\epsilon_2}} + \left(\frac{y}{a_y}\right)^{\frac{2}{\epsilon_2}}\right)}^{\frac{\epsilon_2} {\epsilon_1}} + \left(\frac{z}{a_z}\right)^{\frac{2}{\epsilon_1}}
\end{align}

\noindent where $\textbf{x} = [x, y, z]^T \in \mathbb{R}^3$ is a point or surface vector defined in the superquadric frame. The exponents ($\epsilon_1, \epsilon_2$) produce a variety of convex shapes and describe the shaping characteristics. 

We can fully parameterize a superquadric placed on the table with parameters $\beta = [{\epsilon_1, \epsilon_2, a_x, a_y, a_z}]$ and $g=[x_0, y_0, \theta_0]$ is the initial 2D pose of the object. We aim to robustly optimize the superquadric parameters of the noisy partial point cloud obtained from initial visual observations $o^{V}$. For this, Expectation Maximization is utilized based on the work of \cite{liu2022robust}, which requires casting the estimation problem as a Bayesian inference problem. 

\subsubsection{Estimation of parameters of superquadric}
A Gaussian centroid $\mathbf{\mu} \in \mathcal{S}_{\beta}$, randomly sampled on the superquadric surface $\mathcal{S}_{\beta} \subset \mathbb{R}^3$ and parameterized by $\beta$, according to the uniform density function:
\begin{equation} 
p(\mathbf{\mu}) = \frac{1}{A_{\beta}}, \qquad A_\beta = \int\limits_{\mathcal{S}_\beta} 1d\mathcal{S}
\label{eq:sq_sample}
\end{equation} 
\noindent where $A_\beta$ is the area of superquadric. An observation $\mathbf{x}$ from a Gaussian-uniform model is created in $\mathbb{R}^3$ with a probability density function:

\begin{equation} 
p(\mathbf{x} | \mathbf{\mu}) = w_0p_0(\mathbf{x}) + (1-w_0)\mathcal{N}(\mathbf{x} | \mathbf{\mu}, \Sigma) 
\label{eq:shape_main}
\end{equation} 

\noindent where $\mathcal{N}(.| \mathbf{\mu}, \Sigma)$ denotes the density function of a Gaussian distribution parameterized by ($\mu, \Sigma$). Noise is assumed to be isotropic, i.e. $\Sigma = \sigma^2I$ and $w_0 \in [0, 1]$ represents the probability that a point is sampled from an outlier distribution - $p_0(\mathbf{x}) = 1/V$, with $V$ encapsulating the volume of the interaction area. Eq. \ref{eq:shape_main} is simplified by introducing a latent discrete random variable $\gamma$ that serves as an indicator of the membership of $\mathbf{x}$. When $\gamma=0$ $\mathbf{x}$ is sampled from the uniform outlier component, while, $\gamma=1$ indicates that $\mathbf{x}$ is generated from the Gaussian inlier component. This reformulation results in the following:

\begin{align}
    p(\mathbf{x} | \mathbf{\mu}, \gamma) &= p_0(\mathbf{x})^{1-\gamma}\mathcal{N}(\mathbf{x} | \mu, \Sigma)^\gamma \\
    \gamma \sim p(\gamma) &= Bernoulli(1-w_0)
\end{align}

Given a set of points from the point cloud $\mathcal{PC}_{t_0} = {\mathbf{x}_i \in \mathbb{R}^3 |i = 1, 2, .. N}$ (obtained from initial RGB-D image), the parameters of the superquadric surface can be estimated by minimizing the negative log-likelihood function:

\begin{equation}
    l(\beta, \sigma^2) = \sum_{i=1}^{N} \gamma_i \left(\frac{||\mathbf{x}_i - \mu_i||{^2}_2}{2\sigma^2} -log c\right) + Nlog(A_\beta)
    \label{eq:shape_mle_problem}
\end{equation}

\noindent where $c$ is the normalization constant of the Gaussian distribution. The authors in \cite{liu2022robust} employed a novel Expectation Maximization coupled with a Switching (EMS) approach to solve the MLE problem and overcome the local optimality in Eq.\ref{eq:shape_mle_problem}. The switching approach generates similar candidate superquadrics which are then replaced when the EM algorithm gets stuck in the local minima. In our case, along with the general generation of superquadric candidates as in \cite{liu2022robust}, similar candidates due to nonlinear deformation terms were also generated. The EMS approach first estimates the Gaussian centroids given the current estimation of the superquadric parameters. 
\begin{equation}
    \hat{\mu}_i = \underset{\mu_i \in \mathcal{S}_{\beta}}{\mathrm{argmin}} 
 p(\mu_i|x_i) \sim  \underset{\mu_i \in \mathcal{S}_{\beta}}{\mathrm{argmin}} ||x_i - \mu_i||^2
 \label{eq:shape_mu}
\end{equation}

With the current estimate of $\mu_i$, the expectation of the posterior probability of $x_i$ being an inlier is inferred via the Bayes' rule: 

\begin{equation}
    E(\gamma_i=1|x_i, \hat{\mu}_i) = \frac{\mathcal{N}(x_i |\hat{\mu}_{i}, \sigma^2I)}{\mathcal{N}(x_i|\hat{\mu}_i, \sigma^2I) + \frac{w_op_o(x_i)}{1-w_o}}
\end{equation}

Finally, the parameters $\beta, \sigma$ are optimized via M-Step by substituting the posterior estimates $\hat{\mu}_{i}$ and $\gamma_i$ in Eq. \ref{eq:shape_mle_problem}. The EM step is applied iteratively until the change in parameters $\beta, \sigma$ is less than the threshold $(0.001)$. In robotic settings, it is common to have only a partial point cloud of the object available initially. To improve the shape recovery, we employ three fixed views to capture a detailed and comprehensive point cloud of the object, as shown in Fig. \ref{fig:problem_setup}. Additionally, the inferred superquadric parameter is used to sample the boundary points and determine the push action affordance/parameters. 

\subsection{Cross Modal Gaussian Process Model: CM-GP}
\label{sec:cmgp}
We use the Gaussian process regression \cite{seeger2004gaussian} to model the intricate relationship between the shape parameters $\mathcal{S}_\beta$ and an approximate initial estimate of the prior distribution over the mass as:
\begin{equation}
    p(\phi_{t_0}|S_\beta, X, y) = \mathcal{N}(\phi_{t_0}|\mu_{\phi_{t_0}}, \Sigma_{\phi_{t_0}})
\end{equation}

where the $\mu_\phi, \Sigma_\phi$ are computed using the standard Gaussian process optimization on observed dataset $\mathcal{D}$ comprising of $n$ points, with $X=[\mathcal{S}^1_\beta, .., \mathcal{S}^n_\beta]$ and $y=[\phi^1_{t_H}, .., \phi^n_{t_H}]$. We leveraged a combination of Constant and Radial Basis Function Kernel structure to capture both local and global fluctuations in shape and mass relationship. The prior distribution is utilized within the dual filtering setup explained in the following section. Furthermore, details on the iterative accumulation of points in the dataset $\mathcal{D}$ are presented in Section \ref{sec:surprise}.

\subsection{Dual Filtering}
\label{sec:df}
We represent the belief about the current state of the object $s_t$ with a distribution conditioned on previous actions $a_{1:t}$ and observations $o_{1:t}$. This distribution is denoted as the belief of the state $bel(s_t)$  
\begin{equation}
     bel(s_t) = p(s_t|o_{1:t}, a_{1:t}) = \frac{p(o_t|s_t,o_{1:t-1}, a_{1:t})p(s_t|o_{1:t-1}, a_{1:t})}{p(o_t|o_{1:t-1},a_{1:t})}
     \label{eq:belief_state}
 \end{equation}
One prominent approach to computing the belief tractably is to employ Recursive Bayesian Filters which follow the structure:
\begin{align}
    bel(s_t) = p(s_t|o_{1:t}, a_{1:t}) = \eta p(o_t|s_t,a_t)\overline{bel}(s_t)\\
    \overline{bel}(s_t) = \int_{}^{}p(s_t|s_{t-1}, a_{t-1})bel(s_{t-1})ds_{t-1} 
    \label{eq:bayes_filter}
\end{align}
Kalman Filters are a common choice of Bayesian Filtering which is optimal in linear systems and can be extended in non-linear cases through various approaches. Two key aspects of Bayesian filtering are the representation of the state process model in the form of $p(s_t|s_{t-1}, a_{t-1})$ and an observation likelihood model that relates the states to the observations $p(o_t|s_t)$. In our problem, we use the analytical model of pushing \cite{lynch1992manipulation, mason_pushing} to formulate the process and the observation model. Furthermore, the pose of the object is intricately dependent on the parameters, and straightforward combined (joint) filtering for pose and parameter does not perform well \cite{dutta2023push}. Therefore, we utilize a dual filter design, exploiting the dependency among the states for consistent filtering and inferring the parameters of the object.  

We briefly present our dual filter approach \cite{dutta2023push} for completeness. For the dual filter formulation, we explicitly represent the state $s_t$ by the joint distribution of $\psi_t$ and $\phi_t$, via Multivariate Gaussian distribution: 
\begin{equation}
     bel(\psi_t, \phi_t) \sim \mathcal{N}(\psi_t, \phi_t | \mu_t , \Sigma_t)
     \label{eq:joint_distribution}
 \end{equation}
with statistics $\mu_t \in \mathbb{R}^4$ and $\Sigma_t \in \mathbb{R}^{4\times4}$ as 
\begin{align}
    \mu_t = \begin{pmatrix}
\mu_{\psi_t} \\
\mu_{\phi_t}
\end{pmatrix}, \quad \Sigma_t = \begin{pmatrix}
\Sigma_{\psi_t} & \Sigma_{{\psi_t\phi_t}}\\
\Sigma_{{\phi_t\psi}_t} & \Sigma_{\phi_t}
\end{pmatrix} \quad
\end{align}
The dual filter as shown in Fig.\ref{fig:framework}(b) follows, the structure of a Kalman Filter with a \textit{prediction step} and an \textit{update step} with following modifications: 

\subsubsection{Prediction Step}
In the prediction step, the next step joint belief is predicted given the prior belief and the actions. The object parameters are real physical quantities with some physical constraints (for, e.g. $m > 0$). However, simply constraining the sigma points $\chi^{UT}$ in the UKF approach does not preserve the true variance of the Gaussian distribution \cite{ut_constraints}. Therefore, we performed constrained Monte Carlo sigma point sampling to preserve the physical constraints and the variance of the Gaussian distribution. We employ the sampling method \cite{mc_method} to sample $C$ sigma points in the joint distribution $bel(\psi_{t-1}, \phi_{t-1})$ instead of using standard Unscented Transform points: 
\begin{equation}
     \chi^{[i]}_{t-1} = \mu_{t-1} + \epsilon^{[i]} \sqrt{\Sigma_{t-1}} 
     \label{eq:mc_sampling}
\end{equation}
where, $i \in 1..C$ and $\chi_{t-1} = [\chi_{\psi_{t-1}}, \chi_{\phi_{t-1}}] \in \mathbb{R}^{C\times4}$ with  an associated weight $w_t^{[i]} = 1/C$ and $\epsilon^{[i]} \sim \mathcal{N}(0, 1)$. We set $C=100$ for all of our experiments. The sigma points are filtered on the basis of whether they satisfy the physical constraints and pass through the process and observation models. However, the invalid sigma points are also retained and reintroduced to preserve the uncertainty of the distribution. 

The analytical process model of quasi-static pushing predicts the movement of the object given the previous estimates of the pose sigma points $\chi_{\psi_{t-1}}$, the robot action $a_t$, the normal surface associated at the contact point $n$ estimated from the superquadrics, as well as the estimates of the mass sigma points ${\chi}_{\phi_{t-1}}$, surface friction $\mu$ and friction between the object and the robot $\mu_r$.  The friction of the surface $\mu$ and the friction between the object and the robot are experimentally determined and are assumed to be known in this work. Interested readers are referred to \cite{lynch1992manipulation, kloss2022combining} for details on the analytical model of pushing.
\vspace{-5pt}
\begin{flalign}
    \overline{\chi}_{\psi_t} &\longleftarrow AnalyticalProcess(\chi_{\psi_{t-1}}, {\chi}_{\phi_{t-1}}, a_t) \\
    \overline{\chi}_{\phi_t} &= {\chi}_{\phi_{t-1}}
     \label{eq:process_model}
\end{flalign} 
The diagonal process noise $Q_t \in \mathbb{R}^{C\times3}$ is for a time-varying pose which is tuned manually. The predicted next step sigma points $\overline{\chi}_t$, along with the process noise $Q_t$ are utilized to compute the expected Gaussian belief $\overline{bel}(\psi_t, \phi_t)$ as 
\begin{flalign}
    \overline{\chi}^{[i]}_{\psi_t} = \overline{\chi}^{[i]}_{\psi_t} &+ \epsilon^{[i]} \sqrt{Q_t^{[i]}} \\
    \overline{\mu}_t = \sum_{i=0}^{C}w_t^{[i]} \overline{\chi}_t
	& \quad \overline{\Sigma}_t = \sum_{i=0}^{C}w_t^{[i]}(\overline{\chi}^{[i]}_t-\overline{\mu}_t)(\overline{\chi}^{[i]}_t-\overline{\mu}_t)^T
     \label{eq:predicted_belief}
\end{flalign} 
where, $i \in 1.. C$ and $\overline\chi_{t} = [\overline\chi_{\psi_{t}}, \overline\chi_{\phi_{t}}]$

\subsubsection{Update Step}
We recompute the constrained Monte Carlo sigma point $\overline{\chi}^{'}_{\phi_t}$ sampling on the predicted belief $\overline{bel}(\psi_t, \phi_t)$ to incorporate the process noise. The dual filter employs a separate update of parameter belief similar to the parameter update presented in \cite{liu_west} and the conditional pose belief update based on the UKF update \cite{prob_rob_book}.  

To update the joint belief, we require an observation model to predict the observation sigma points $\overline{\mathcal{Z}}_t$, which must account for visual and tactile observations. To reduce the complexity of predicting raw RGB-D images, we use the initial segmented point cloud $\mathcal{PC}_{t_0}$ from the shape perception method to transform it using the predicted pose and generate expected RGB-D images using the standard 3D to 2D projective transformation approach \cite{nematollahi2022t3vip} involving the intrinsic and extrinsic values of the camera. This overcomes the generalization problem faced by synthetic visual networks as in \cite{dutta2023push} and can be used for any novel object. For the tactile counterpart, the expected contact force is estimated using the computed analytic movement of the object and the quasi-static limit surface approximation \cite{lynch1992manipulation} for each sigma point. The predicted observation sigma points $\overline{\mathcal{Z}}_t$ are given by:

\begin{flalign}
     \overline{\mathcal{Z}}^{V}_t &= \mathbf{w}(\overline{\chi}^{'}_{{\psi}_t}, \mathcal{PC}_{t_0}) \\
     \overline{\mathcal{Z}}^{T}_t &\longleftarrow  AnalyticalObs(\overline{\chi}^{'}_t)
    \label{eq:observation_model}
\end{flalign}

\noindent where $\mathbf{w}$ is the projective transformation function. 
 RGB-D images $o^{V}_t$ are transformed into grayscale and resized to size $64 \times 64$. They are then flattened and merged with tactile observations $o^{T}_t$ to create $z_t \in \mathbb{R}^{4098}$, which is used for the update process. The diagonal observation noise $R_t$ is composed of 2 values $\sigma^{{obs}^{V}}$ and $\sigma^{{obs}^{T}}$, tuned manually.
\begin{equation}
R_t = diag [\underbrace{\sigma^{{obs}^{V}}_t, \cdots \sigma^{{obs}^{V}}}_{4096} , \sigma^{{obs}^{T}}, \sigma^{{obs}^{T}}]
\end{equation}


\subsubsection*{Parameter Update}
We update the weights based on the likelihood of the observation sigma points $\overline{\mathcal{Z}}_t = [\overline{\mathcal{Z}}^{T}_t, \overline{\mathcal{Z}}^{V}_t] $ in the observation distribution $\sim \mathcal{N}(.|z_t, R_t)$
\begin{flalign}
    w_t^{j} &=  w_t^{j}e^{\big(-\frac{1}{2}(\overline{\mathcal{Z}}^{j}_t - z_t) R^{-1}(\overline{\mathcal{Z}}^{j}_t - z_t)^T)}
     \label{eq:parameter_update}
\end{flalign}
where $j \in 1, .. C$. The updated parameter belief $bel(\phi_t)$ is recomputed  via a Gaussian Smooth Kernel \cite{liu_west} method after normalizing the updated weights:
\begin{flalign}
    \mu_{\phi_t} &= \sum_{i=0}^{C}w_t^{[i]} \overline{\chi}^{'}_{\phi_t};  \quad m^{[i]}_{\phi_t} = a\overline{\chi}^{'}_{{\phi}_t}+(1-a)\mu_{\phi_t} \\
	\Sigma_{\phi_t} &= h^2\sum_{i=0}^{C}w_t^{[i]}m^{[i]}_{\phi_t}-\mu_{\phi_t}
\end{flalign}
where $a$ and $h=\sqrt{1-a^2}$ are shrinkage values of the kernels that are user-defined and set to 0.01, and $m$ are the kernel locations.

\subsubsection*{Pose Update}
We make use of the dependence of the pose on the parameters to compute the conditional pose distribution $bel(\psi_t | \phi_t) \sim \mathcal{N}(\psi_t|\mu_{\psi_t|\phi_t}, \Sigma_{\psi_t|\phi_t})$ using the Multivariate Gaussian Theorem \cite{mult_gauss}
\begin{flalign}
    \mu_{\psi_t|\phi_t} &= \psi_t + \Sigma_{\psi_t\phi_t}\Sigma^{-1}_{\phi}(\phi_t - \mu_{\phi_t})\\
    \Sigma_{\psi_t|\phi_t} &= \Sigma_{\psi_t} -  \Sigma_{\psi_t\phi_t}\Sigma^{-1}_{\phi_t}\Sigma_{\phi_t\psi_t}
    \label{eq:cond_pose_distribution}
\end{flalign}

For the update of the conditional pose, the standard Unscented Kalman Filter (UKF) \cite{prob_rob_book} is used on the predicted conditional pose distribution $\overline{bel}(\psi_t | \phi_t = \mu_{\phi_t})$ using Eq.\ref{eq:cond_pose_distribution}. The $\mu_{\phi_t}$ of the updated parameter belief is utilised with predicted pose sigma points $\overline{\chi}^{UT}_{\psi_t}$ to obtain the predicted observation sigma points $\overline{\mathcal{Z}}'_t$. After the conditional pose update, the posterior joint is computed as
\begin{equation}
     bel(\psi_t, \phi_t) = bel(\psi_t|\phi_t)bel(\phi_t)
     \label{eq:belief_joint_update}
 \end{equation}
Note that the cross-covariance matrices $\Sigma_{\psi_t\phi_t}, \Sigma_{\phi_t\psi_t}$ are not updated through the dual update step and are kept constant. The updated posterior joint belief is used as a prior to filter, then as the next time step, until the complete sequence is filtered $t=t_H$.  In addition, the initial belief of the parameter $bel(\phi_{t_0}) \sim \mathcal{N}(\phi_{t_0}|\mu_{\phi_{t_0}}, \Sigma_{\phi_{t_0}})$ estimated from the \textit{CM-GP}.

\subsection{Iterative Training of Cross-Modal GP: Surprise}
\label{sec:surprise}
It is well known that the computational complexity of Gaussian process regression methods increases with the number of data points, scaling as $\mathcal{O}(N^3)$. To address this issue, various sparsification techniques have been suggested \cite{schreiter2016efficient}. In this study, we leverage the predictive perception framework to introduce a new form of sparsity. Our goal is to incorporate only the training points that exhibit significant differences compared to the others. To achieve this autonomously and in an unsupervised manner, we propose a `surprise' metric defined as: 

\begin{equation} 
\text{surprise} = (\mu_{\phi_{t_0}} - \mu_{\phi_{t_H}})\Sigma_{\phi_{t_0}}(\mu_{\phi_{t_0}} - \mu_{\phi_{t_H}})^T 
\end{equation} 

This metric represents the negative log-likelihood of the posterior mean parameter estimated by the dual filtering step of the object parameter relative to the prior distribution estimated by \textit{CM-GR}. Hence, a high `surprise' value indicates a data point with shape-to-mass characteristics that deviate significantly from what the current cross-modal model anticipates, prompting its inclusion in the dataset $\mathcal{D}$. Conversely, a low `surprise' value implies that the cross-modal prediction was sufficiently accurate, and the shape-to-mass data point does not need to be added to the dataset. This approach promotes sparsity in the Gaussian process regression model, facilitating efficient long-term learning and prior estimation. We introduce a threshold parameter $\alpha$ on the computed surprise value to control the addition of data points to the dataset, which is fine-tuned for optimal performance.

\section{Experiments}
\label{sec:results}

\begin{figure*}[tbh!]
    \centering
    \includegraphics[width = 0.85\textwidth]{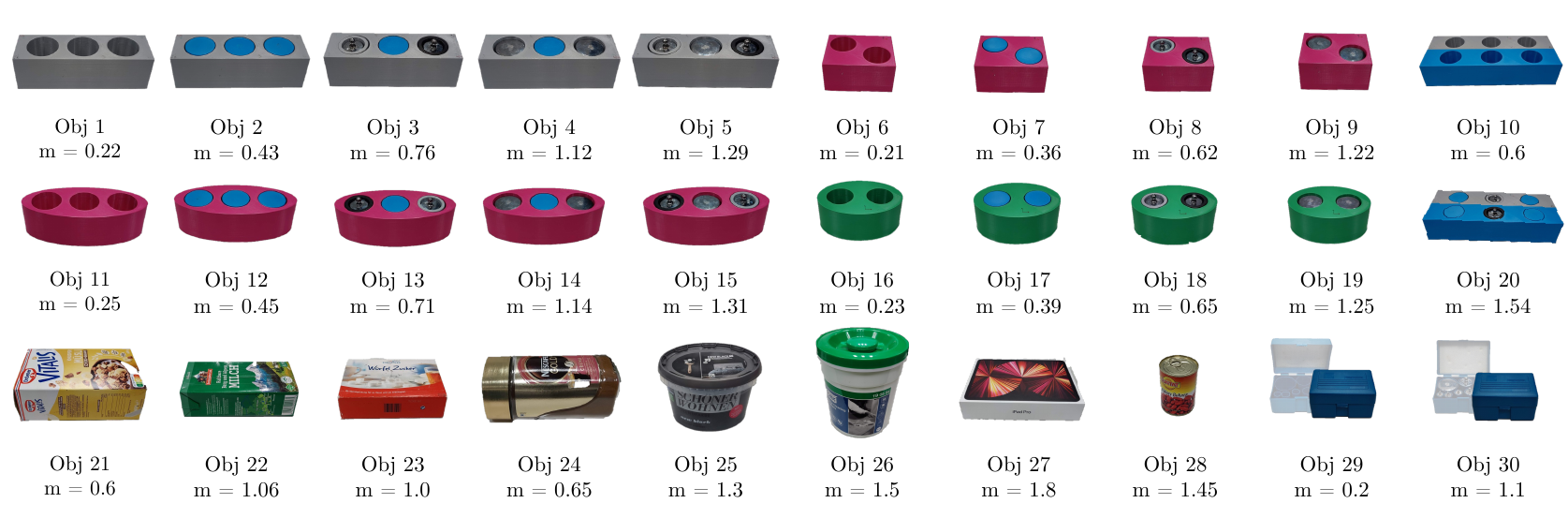}
    \caption{Experimental Object List. Object name followed by measured GT mass in kg}
    \label{fig:experimentalobj}
\end{figure*}

\begin{figure}[tbh!]
    \centering
    \includegraphics[width = \columnwidth]{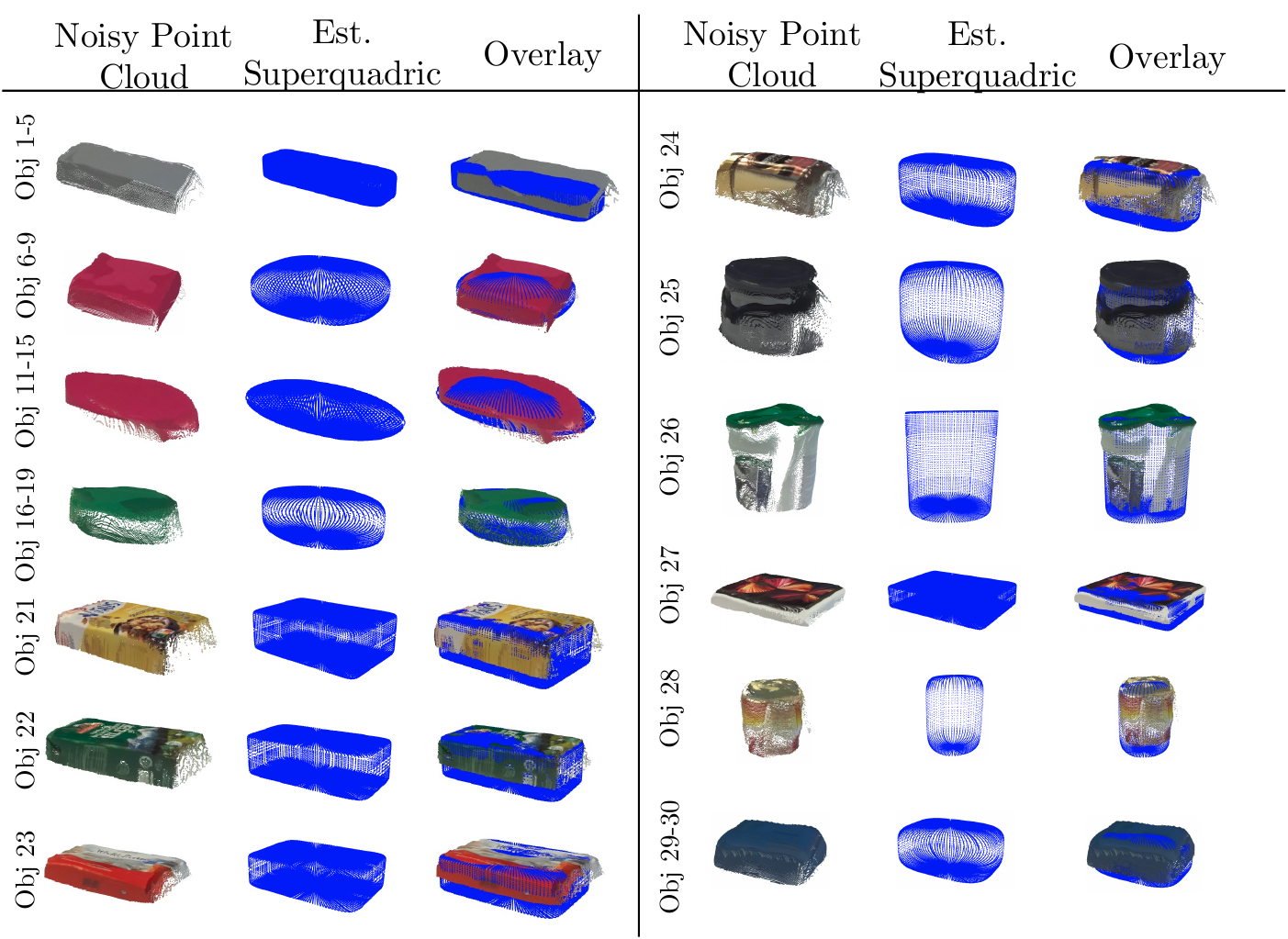}
    \caption{Qualitative shape estimation results on the distinct shapes}
    \label{fig:qual_shape}
\end{figure}

\begin{figure*}[tbh!]
    \centering
    \includegraphics[width=\textwidth]{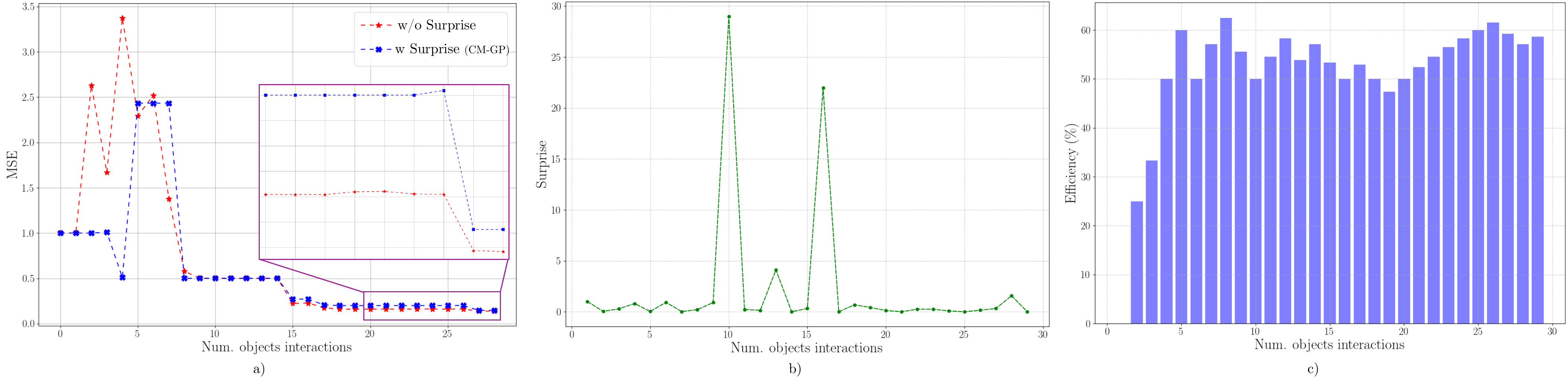}
    \caption{a) Mean squared error value of the predicted prior mass values w.r.t GT mass. The proposed $CM-GP$ (\textit{with surprise}) utilises fewer data points, with a similar and consistent performance over  \textit{without surprise} model where all estimated shape and mass values are used for training. b) Evaluation of the estimated `surprise' value during each object-robot interactions c) Efficiency of the proposed $CM-GP$ model in terms of $\%$ less data used at each iteration. A higher value signifies a more efficient Gaussian process regression model.}
    \label{fig:resultscomb}
\end{figure*}

In this section, we describe the experimental setup and the results obtained from the proposed framework. A real robot set-up was used to validate the suggested approach and compare it with the baseline method. The robotic setup depicted in Fig. \ref{fig:problem_setup} comprises a Universal Robots UR5 equipped with a Robotiq two-finger gripper and a Franka Emika Panda robotic manipulator, as illustrated in Fig. \ref{fig:problem_setup}. Tactile sensors known as Contactile sensors \cite{contactile} were attached to the outer surface of the Robotiq gripper finger pad, while a Zed2i stereo camera \cite{zed} was firmly attached to the Panda arm. The maximum speed allowed for the UR5 and Panda robots was limited to 25 $mm/s$ for safety reasons.

To validate our proposed method, we developed 20 configurable 3D-printed objects of different shapes. We varied the mass of each of these objects by adding weights ranging from $0.07$ to $1$ kg. In addition, we selected 10 everyday objects with diverse shapes and mass values. The list of experimental objects can be found in Fig. \ref{fig:experimentalobj}. Each interaction between the object and the robot involved manually positioning the object in the center of the tabletop surface, after which the proposed framework was autonomously executed. For each object, we also measured the $GT$ mass using a digital scale for evaluation.

\begin{figure}[tbh!]
    \centering
    \includegraphics[width=0.85\columnwidth]{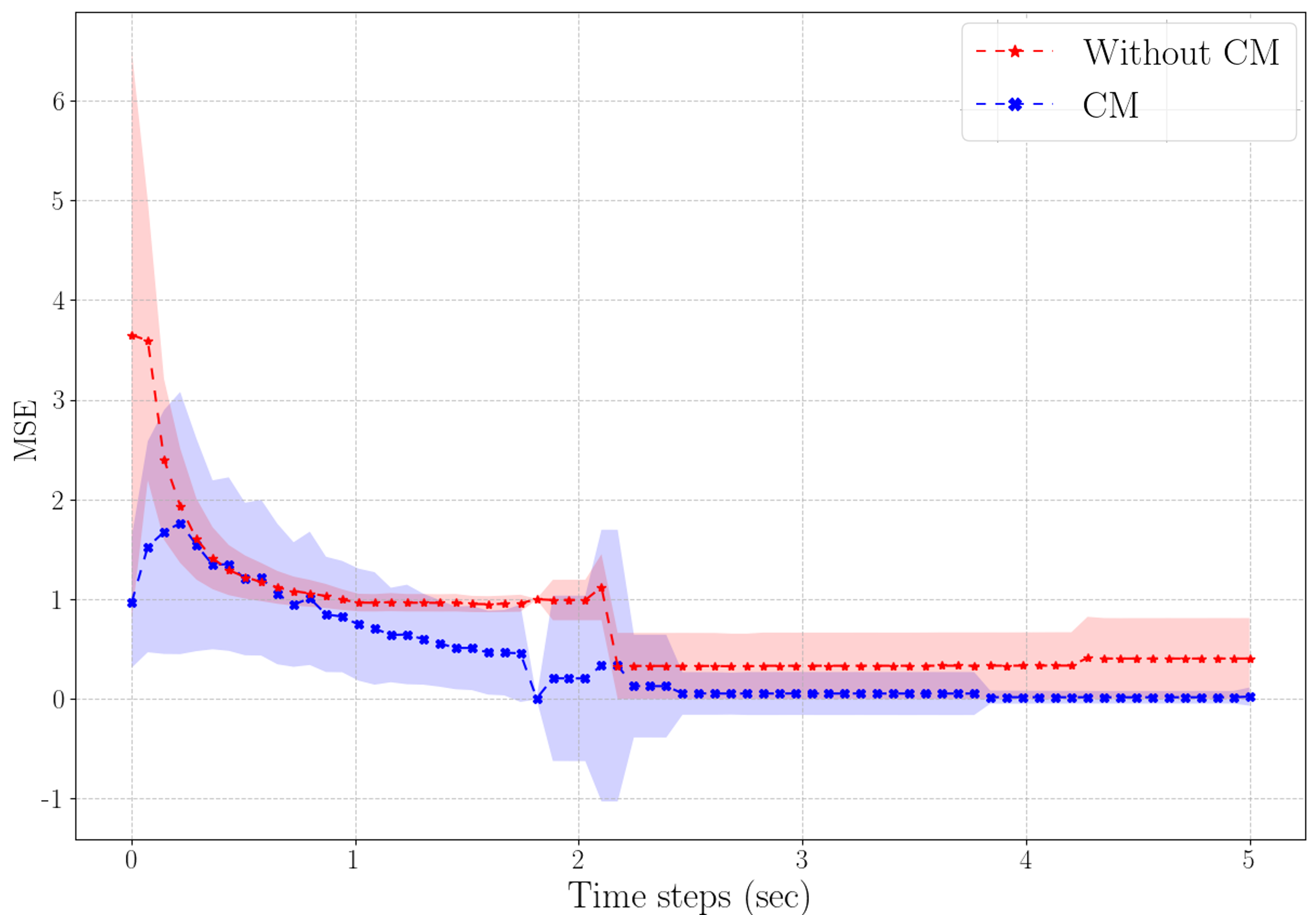}
    \caption{Inference result of the dual filter of mass estimation with proposed cross-modal prior (in blue) and without cross-modal prior (in red)}
    \label{fig:ddfmse}
\end{figure}

The qualitative results of the shape of each object, estimated by the superquadric method, are shown in Fig. \ref{fig:qual_shape}. This approach successfully captures the various shapes of objects with precision, making it suitable for use by $CM-GP$. We present the results of the parameter inference of the dual filtering approach using the cross-modal prior ($CM$) in Fig. \ref{fig:ddfmse}. The figure illustrates the evolution of the mean squared error ($MSE$) between the estimated mass and the value of the GT mass during the push interaction that lasts $5 sec$. The inference process was carried out iteratively on successive objects coupled with the training of \textit{CM-GP}. Furthermore, we compared the performance of parameter inference when the visual prior of the cross-modal approach was not applied, opting instead for a single mean value of $\mu_{\phi_{t_0}} = 0.5$ and $\sigma_{\phi_{t_0}} = 0.25^2$ as the prior distribution for the inference step (\textit{without CM}). The results indicate that with the cross-modal approach, the estimation converges 1-2 seconds more efficiently with reduced errors.

We also demonstrate the predictive capability of the cross-modal Gaussian process model ($CM-GP$) following each interaction between objects and the robot, as shown in Fig. \ref{fig:resultscomb}(a). The plot presents the mean square error between the predicted mass and $GT$ mass values of all objects regressed through the iteratively trained \textit{CM-GP} model. The result shows that the model is capable of predicting a reasonable mass estimate ($MSE < 0.5$) after 15 interactions. As a supporting result, the computed 'surprise' values during object interactions are depicted in Fig. \ref{fig:resultscomb}(b). A higher 'surprise' value $(>=1)$ leads to the incorporation of shape-mass values in the dataset $\mathcal{D}$.

\begin{figure}[tbh!]
    \centering
    \includegraphics[width=0.85\columnwidth]{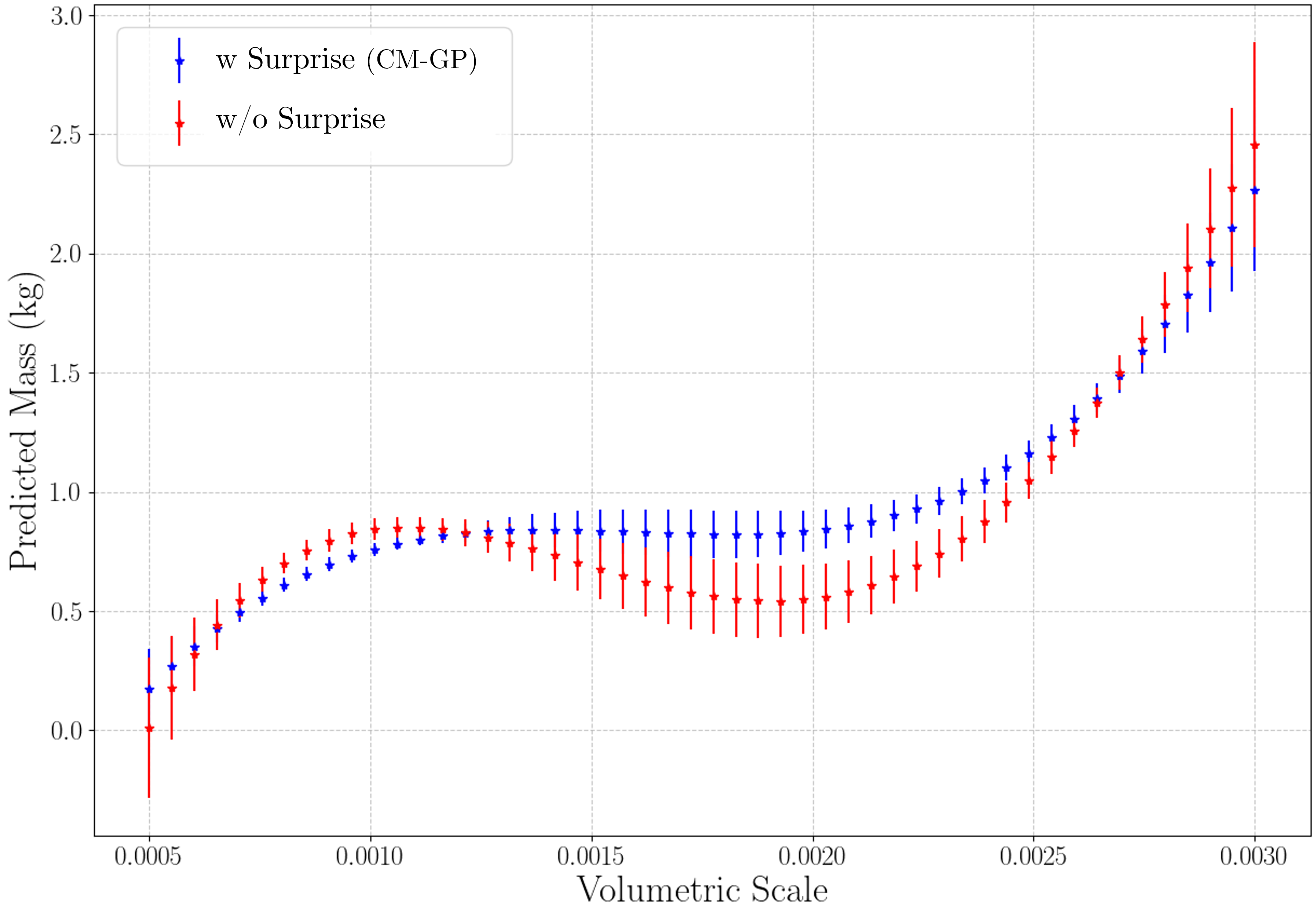}
    \caption{Comparison of the predicted mass value of $CM-GP$ model (\textit{with surprise}) vs \textit{without surprise} approach for uniform shape change depicted by gradually increasing volumetric scale $=1/(\sqrt{a^2_x + a^2_y + a^2_z}$ with $\epsilon_1=\epsilon_2 = 0.5$  }
    \label{fig:cmgp}
\end{figure}

We also present a comparison with a Gaussian Process Regression model (\textit{without surprise}) in which the proposed surprise formulation was not used and all shape mass values were used in training. It is evident that with the use of our proposed `surprise' formulation and sparse data points, the cross-modal performance remains comparable to that achieved when using all estimated shapes and mass points. This is further highlighted in Fig. \ref{fig:cmgp}, which presents regressed uniformly sampled shape parameters and the corresponding prediction of the mass value after 30 interactions. Furthermore, we present the efficiency of the proposed $CM-GP$ with that of \textit{without surprise} in Fig. \ref{fig:resultscomb}(c). The efficiency metric is calculated as the percentage of shape-to-mass data points utilized at each interaction. We can observe that during the subsequent object-robot interaction, at most $50\%$ of the shape-mass values are used to train the GP model, leading to improved computational efficiency. 

\section{Discussion \& Conclusion}
In this study, we introduce a novel visuo-tactile based predictive cross-modal perception framework for interactive object exploration in robotics. Our experimental results illustrate the improvement of parameter estimation (mass) by integrating a visual prior provided by our proposed cross-modal Gaussian process model. Moreover, by employing our interactive non-prehensile pushing and dual-filter approach, we introduce a novel `surprise' formulation. This formulation iteratively refines the cross-modal Gaussian process model $CM-GP$, leading to a substantial reduction of around 50\% in the required data points. 

We performed real-robotic experiments in which the objects were selected to critically evaluate our proposed approach, in which similar shapes had varying mass values, requiring not only a consistent prior value but associated uncertainty for accurate filtering. The proposed approach $CM-GP$ seamlessly integrates and improves our previous dual filtering method for the estimation/inference of object parameters. 

In the future, we plan to investigate more complex cross-modal capabilities that can predict various object characteristics such as surface friction based on visual cues like colour and reflectivity. Furthermore, we noticed that the effectiveness of the $CM-GP$ model was influenced by the order in which objects were used. This is not surprising given the iterative training process, but we aim to thoroughly assess various kernels that evolve over iterations to mitigate the impact of object sequence.

In conclusion, our work tackles the formidable challenge of visuo-tactile based cross-modal perception in robotic systems. With our proposed predictive approach, robotic systems can autonomously and efficiently explore objects in real-world scenarios without the need for supervision, enabling lifelong cross-modal learning capabilities.

\section*{Acknowledgment}
We would like to thank Prof. Thrishantha Nanayakkara for his insights and comments on the work.

\bibliography{biblio}
\bibliographystyle{IEEEtran}

\end{document}